%% file: sample-sigconf.tex
\documentclass[sigconf]{acmart}
\AtBeginDocument{%
  }

\copyrightyear{2025}
\acmYear{2025}
\setcopyright{acmlicensed}\acmConference[MM '25]{Proceedings of the 33rd ACM International Conference on Multimedia}{October 27--31, 2025}{Dublin, Ireland}
\acmBooktitle{Proceedings of the 33rd ACM International Conference on Multimedia (MM '25), October 27--31, 2025, Dublin, Ireland}
\acmDOI{10.1145/3746027.3758275}
\acmISBN{979-8-4007-2035-2/2025/10}

\usepackage{subfigure}
\usepackage{amsmath}
\usepackage{balance}
\usepackage{pifont}
\usepackage{multirow}
\usepackage{cleveref}
\usepackage{hyperref}

\begin{document}

\title{Investigating Domain Gaps for Indoor 3D Object Detection}

\author{Zijing Zhao}
\email{zijingzhao@stu.pku.edu.cn}
\affiliation{%
  \institution{Wangxuan Institute of Computer Technology, Peking University}
  \city{Beijing}
  \country{China}
}

\author{Zhu Xu}
\email{xuzhu@stu.pku.edu.cn}
\affiliation{%
  \institution{Wangxuan Institute of Computer Technology, Peking University}
  \city{Beijing}
  \country{China}
}

\author{Qingchao Chen}
\email{qingchao.chen@pku.edu.cn}
\affiliation{%
  \institution{National Institute of Health Data Science, Peking University}
  \city{Beijing}
  \country{China}
}

\author{Yuxin Peng}
\email{pengyuxin@pku.edu.cn}
\affiliation{%
  \institution{Wangxuan Institute of Computer Technology, Peking University}
  \city{Beijing}
  \country{China}
}

\author{Yang Liu}
\authornote{Corresponding author}
\email{yangliu@pku.edu.cn}
\affiliation{%
  \institution{Wangxuan Institute of Computer Technology, Peking University}
  \city{Beijing}
  \country{China}
}

\renewcommand{\shortauthors}{Zijing Zhao, Zhu Xu, Qingchao Chen, Yuxin Peng, \& Yang Liu}

\begin{abstract}
  As a fundamental task for indoor scene understanding, 3D object detection has been extensively studied, and the accuracy on indoor point cloud data has been substantially improved. 
  However, existing researches have been conducted on limited datasets, where the training and testing sets share the same distribution. 
  In this paper, we consider the task of adapting indoor 3D object detectors from one dataset to another, presenting a comprehensive benchmark with ScanNet, SUN RGB-D and 3D Front datasets, as well as our newly proposed large-scale datasets ProcTHOR-OD and ProcFront generated by a 3D simulator. 
  Since indoor point cloud datasets are collected and constructed in different ways, the object detectors are likely to overfit to specific factors within each dataset, such as point cloud quality, bounding box layout and instance features. 
  We conduct experiments across datasets on different adaptation scenarios including synthetic-to-real adaptation, point cloud quality adaptation, layout adaptation and instance adaptation, analyzing the impact of different domain gaps on 3D object detectors. 
  We also introduce several approaches to improve adaptation performances, providing baselines for domain adaptive indoor 3D object detection, hoping that future works may propose detectors with stronger generalization ability across domains.
  The benchmark datasets and baseline code are available on our project homepage: \url{https://jeremyzhao1998.github.io/DAVoteNet-release}.
\end{abstract}



\input{figures/teaser}


\maketitle

\input{chapters/1.introduction}
\input{chapters/2.related_work}
\input{chapters/3.benchmark}
\input{chapters/4.Experiment}
\input{chapters/5.conclusion}

\bibliographystyle{ACM-Reference-Format}
\balance
\bibliography{sample-base}

\end{document}

%% file: figures/teaser.tex
\begin{teaserfigure}
    \vspace{-5mm} 
    \begin{center}
        \subfigure[Indoor point clouds visualization in different datasets]{
            \centering
            \includegraphics[width=0.38\textwidth]{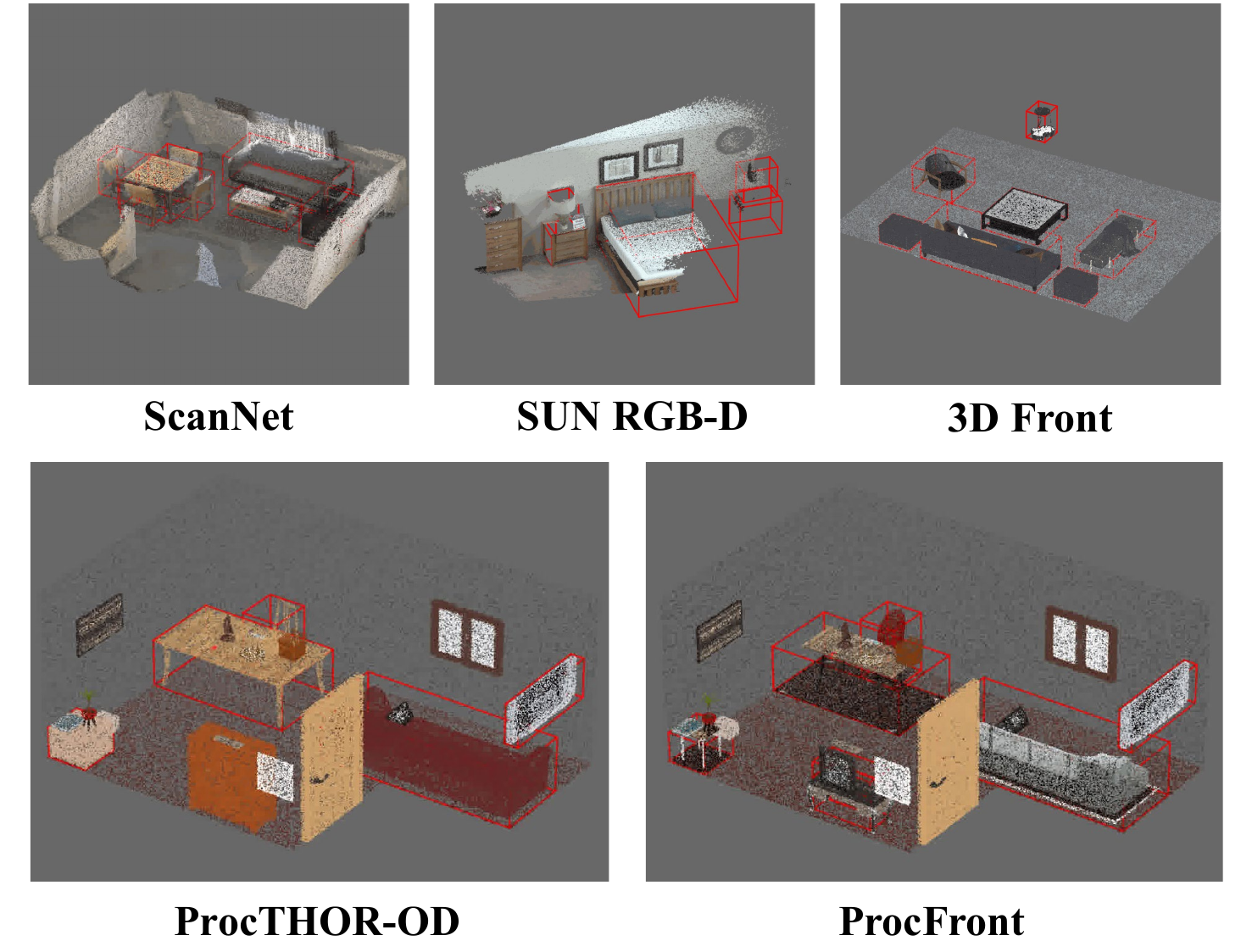}
            \label{fig:teaser-a}
        }
        \hspace{8mm}
        \subfigure[Performance within and across datasets]{
            \centering
            \includegraphics[width=0.42\textwidth]{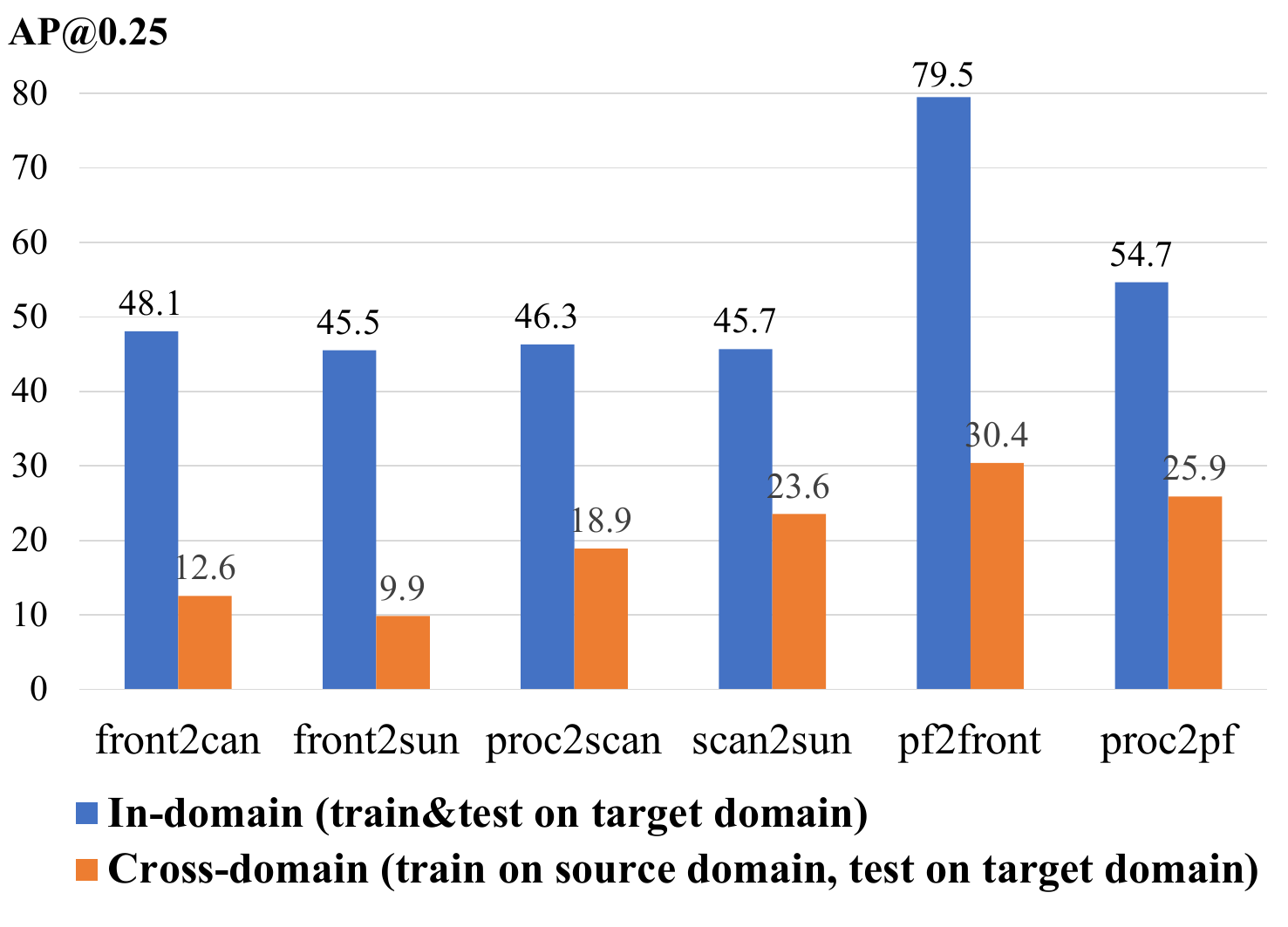}
            \label{fig:teaser-b}
        }
    \end{center}
    \vspace{-5mm} 
    \caption{
        (a) Point clouds in different datasets exhibit significant differences in point cloud quality, bounding box layout and object features.
        (b) mAP within and across datasets. The performance shows a drastic decline in cross-dataset evaluation.
    }
    \Description{}
    \label{fig:teaser}
\end{teaserfigure}

%% file: chapters/1.introduction.tex
\section{Introduction}

As a fundamental task for 3D perception and indoor scene understanding, indoor 3D object detection has been extensively studied.
Most 3D object detectors for indoor scenes are designed for point cloud data due to its adaptability, precision, and richness of information.
Indoor 3D detectors \cite{qi2019deep,pan20213d,shen2023vdetr} have achieved remarkable success in identifying and localizing objects in point clouds.

Though achieving great progress, existing indoor 3D object detection researches are mainly conducted on ScanNet dataset \cite{dai2017scannet} and SUN RGB-D dataset \cite{song2015sun}.
Detector are trained and evaluated within each dataset where the training and testing set share the same data distribution.
As illustrated in \cref{fig:teaser-a}, point clouds in ScanNet dataset are constructed by thorough scans of RGB-D videos, thus with higher quality but relatively fewer scenes.
Point clouds in SUN RGB-D dataset are converted through single RGB-D images, exhibiting large areas of point omission.
In practical applications, detectors trained on data from a specific domain may need to generalize to deployment environments with distributional shift, i.e. domain adaptation problem which has rarely been explored in indoor 3D object detection field.

Domain adaptation for point cloud data has been studied on object classification task \cite{qin2019pointdan,shen2022domain,cardace2023selfdistill}, outdoor LiDAR detection task \cite{wang2020train,liu2024bevuda,chang2024cmda} and indoor semantic segmentation task \cite{ding2022doda}.
However, object detection on indoor point clouds has the following challenges:
(1) Compared with 3D instance classification and outdoor LiDAR detection task, indoor scene point clouds can be constructed through multiple ways, creating more aspects of domain gap factors with larger domain gap to be solved.
(2) Compared with outdoor LiDAR detection, indoor 3D detection requires to distinguish multiple object categories (more than 10 categories v.s. only detecting cars) with higher object diversity, resulting in richer semantics to be learned and adapted.
(3) Localizing objects in indoor 3D scenes faces the challenges of close proximity of objects, sometimes even overlapping bounding boxes (e.g. a chair placed under a table), which may not appear in classification, segmentation and outdoor LiDAR detection tasks.

Only a few studies (e.g., \cite{wang2024ohda}) have explored domain adaptation techniques for indoor 3D object detection, while their focus has been limited to the synthetic-to-real adaptation scenario. 
However, indoor 3D object detection encompasses a broader range of domain variations and practical application scenarios, which are not adequately represented or evaluated by current benchmarks.
To facilitate a more systematic investigation of the various sources of domain shift that affect the performance of indoor 3D object detectors, we introduce a suite of comprehensive benchmarks. 
These benchmarks are constructed from both widely used public datasets and newly developed datasets designed to capture diverse conditions.

As shown in \cref{fig:teaser-a}, we construct our benchmarks using five representative indoor scene datasets.
ScanNet \cite{dai2017scannet} represents high-quality real-world point clouds acquired through dense scanning.
SUN RGB-D \cite{song2015sun} also contains real scenes but is captured with a single RGB-D image per scene, resulting in sparse and incomplete point clouds.
3D Front \cite{fu20213d} provides synthetic indoor layouts with detailed 3D furniture models, but its scale is limited due to manual layout design by human experts.
We further introduce two new datasets: ProcTHOR-OD and ProcFront.
We programmatically generates layouts and places synthetic objects by a simulator platform to produce ProcTHOR-OD, enabling large-scale and diverse scene generation.
ProcFront uses the same layouts as ProcTHOR-OD but integrates instances from 3D Front, allowing controlled analysis of layout and instance domain gaps.

After selecting a consistent subset of instance categories across datasets, we combine the five datasets to construct a suite of domain adaptation benchmarks, covering the following scenarios:
(1) Synthetic-to-real adaptation: Following \cite{wang2024ohda}, we evaluate 3D Front $\rightarrow$ ScanNet and 3D Front $\rightarrow$ SUN RGB-D, and additionally propose ProcTHOR-OD $\rightarrow$ ScanNet as a new synthetic-to-real setting where the source domain has larger data scale.
(2) Point cloud quality adaptation: We test ScanNet $\rightarrow$ SUN RGB-D to assess the performance drop when transferring from high-quality scans to lower-quality, single-frame point clouds.
(3) Layout adaptation: To isolate the impact of layout differences, we evaluate ProcFront $\rightarrow$ 3D Front. The two datasets share the same instance models but differ in scene layouts.
(4) Instance adaptation: To examine domain gaps at the instance level, we evaluate ProcTHOR-OD $\rightarrow$ ProcFront, where the layouts remain the same but object appearances vary.

Under such benchmarks, we conduct experiments on the commonly used VoteNet \cite{qi2019deep} detector.
As shown in \cref{fig:teaser-b}, the detector performs well within each dataset (blue bars, trained and evaluated within target dataset), but cross-dataset evaluation shows a significant performance drop (orange bars, trained by source and evaluated on target dataset).
We also implement several commonly used domain adaptation approaches in other tasks to improve adaptation performances.
These approaches serve as baselines for domain adaptive indoor 3D object detection, hoping that future works may propose detectors or frameworks with stronger generalization ability across domains.
Our project homepage is: \url{https://jeremyzhao1998.github.io/DAVoteNet-release}.

We summarize the main contributions of this paper as follows:
\begin{itemize}
\item We introduce two new datasets, ProcTHOR-OD and ProcFront which are automatically generated in simulation, offering high scalability and controllability, and are well-suited for studying domain adaptation for indoor 3D object detection.
\item We propose a comprehensive suite of  benchmarks by combining ScanNet, SUN RGB-D, 3D Front, and our newly proposed datasets to isolate and evaluate key domain gap factors, including synthetic-to-real shifts, point cloud quality, layout variations, and instance-level differences.
\item We implement and evaluate several domain adaptation strategies for indoor 3D object detection, providing strong baselines for future research.
\end{itemize}

%% file: chapters/2.related_work.tex
\section{Related works}

\input{figures/dataset-vis}

\subsection{Indoor point cloud datasets}

As a universal 3D structure representation, point cloud data has been widely used in 3D indoor scene understanding tasks.
Various 3D indoor scene datasets \cite{armeni20163d, song2015sun, dai2017scannet} have been proposed by multiple construction ways.
SUN RGB-D dataset \cite{song2015sun} converts single RGB-D images into 3D point clouds.
ScanNet dataset \cite{dai2017scannet} collects RGB-D videos through large amount of indoor rooms with rich semantic annotations, and has been the most widely used dataset in indoor 3D object detection and semantic segmentation task.
Simulated datasets like Structured3D \cite{zheng2020structured3d} and 3D Front \cite{fu20213d} are proposed to accomplish large-scale training with relatively low cost. 
However, these datasets are created by human designers and thus still have high construction cost.
ProcTHOR \cite{deitke2022️procthor} offers an indoor scene generation framework with a fully simulated dataset which mainly focus on embodied AI tasks such as navigation.

Despite their contributions to 3D foundation model training\cite{yang20243d,wang2025unipre3d} and downstream tasks including 3D grounding\cite{guo2025ground,li20253dwg,li2025seeground}, VQA\cite{mo2024bridgeqa,luo2025dspnet} and 3D content generation\cite{qin2025gen,ye2024maan,ye2024relscene,deng2025global,liang2025sinf}, existing datasets remain limited in scale, quality, and label diversity.
They also assume identical distributions between training and testing sets, whereas this paper focuses on domain adaptation for indoor 3D object detection.
To this end, we introduce synthetic ProcTHOR-OD and ProcFront datasets, and combine them with existing datasets to establish new domain adaptation benchmarks.

\subsection{Indoor 3D object detection on Point Clouds}

Indoor 3D object detection is a fundamental task in indoor scene understanding, serving as the upstream task for 3D visual grounding, question answering and navigation.
3D-SIS \cite{hou20193d} back projects the 2D feature vector onto the associated voxel in the 3D grid to achieve 3D detection and segmentation.
VoteNet by \cite{qi2019deep} adopt deep hough voting strategy to train the model to group points for detection proposal generation.
In recent years, transformer architectures are also introduced to detect objects in point clouds \cite{shen2023vdetr,yang2023swin3d,zhu2023spgroup3d}.
V-DETR \cite{shen2023vdetr} equips the detection transformer with 3D Vertex Relative Position Encoding which computes each point’s encoding relative to the box vertices predicted by queries to enforce locality.

Though achieving great success, these studies all focus on performances where the training and testing set share the same data distribution.
The domain adaptation ability of indoor 3D object detectors has not yet been explored.
In this paper, we mainly conduct experiments on VoteNet \cite{qi2019deep} due to its popular use among indoor scene understanding tasks and its lightweight architecture.

\subsection{Domain adaptation for point clouds}

Domain adaptation has been extensively studied in 2D computer vision \cite{pei2024evidential,deng2025groto,chen2018raan,pei2023utr,zhang2025colearnpp,xu2022cda,chen2020sffuda} and downstream tasks such as detection\cite{zhao2023masked,pan2025enhance}, segmentation\cite{luo2025sdgpa,wang2025pulling} and retrieval\cite{chen2021mindthegap,liu2021adaptive,luo2025gvd,luo2025graph}. 
Prior works on 3D domain adaptation has primarily focused on classification, segmentation, and LiDAR point cloud understanding. 
PointDA \cite{qin2019pointdan} first introduced a domain adaptation method for 3D classification by reconstructing local point features. 
Subsequent studies \cite{shen2022domain, cardace2023selfdistill} proposed global feature reconstruction and self-training to improve adaptation performance. 
For semantic segmentation, \cite{ding2022doda} explored synthetic-to-real adaptation via data augmentation and mixing.

LiDAR-based detection in autonomous driving scenes has a close relationship with indoor point cloud object detection.
The pioneer work \cite{wang2020train} evaluate cross-dataset performance of LiDAR detectors, followed by multiple techniques \cite{hu2023density, liu2024bevuda, chang2024cmda} that enhance domain adaptation performance by self-training, point completion, and pseudo-label refinement. 
Nonetheless, indoor 3D detection poses distinct challenges: 
(1) unlike audonomous driving datasets that are all LiDAR-generated, indoor point clouds arise from diverse sources, leading to more complex and varied domain gaps; 
(2) LiDAR point cloud detection typically focuses on the car category, whereas indoor detection involves a wider range of object classes (more than 10) with richer semantics.
In this paper, we investigate the domain gaps for indoor 3D object detection task, providing the first comprehensive benchmark and several baselines.

%% file: figures/dataset-vis.tex
\begin{figure*}
    \centering
    \includegraphics[width=0.95\textwidth]{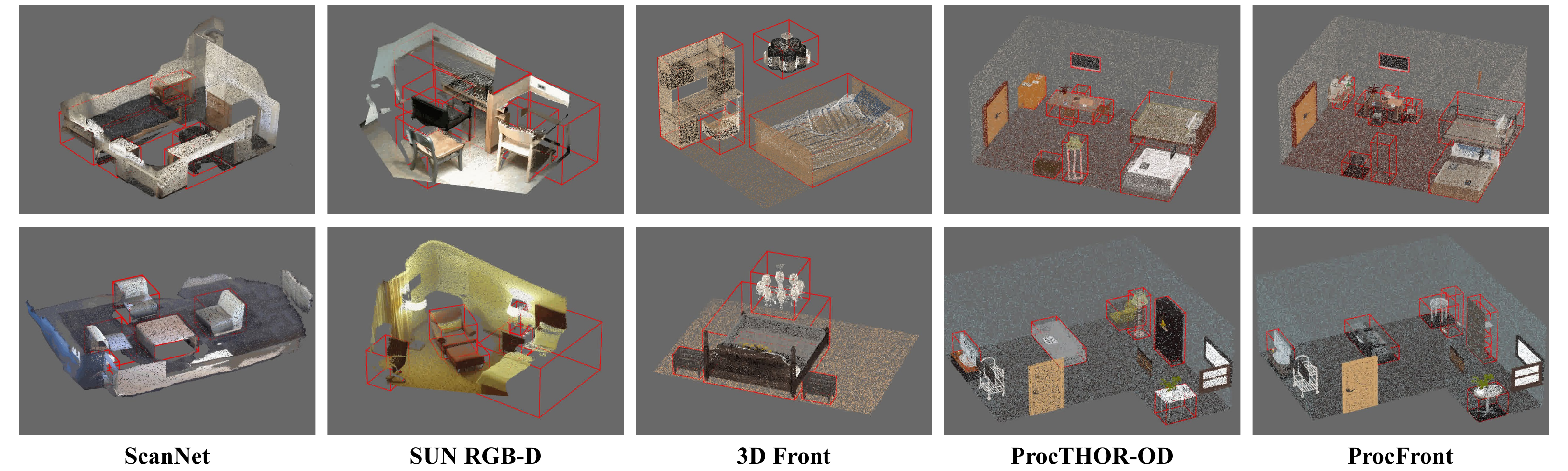}
    \caption{
        Visualization for typical scenes in the selected datasets of our proposed benchmarks.
    }
    \label{fig:dataset-vis} 
    \Description{}
\end{figure*}

%% file: chapters/3.benchmark.tex
\section{Datasets and benchmarks}

To build the domain adaptation benchmarks for indoor 3D object detection, in this section, we review existing indoor 3D object detection datasets in \cref{sec:existing_dataset} and introduce our newly proposed datasets ProcTHOR-OD and ProcFront in \cref{sec:sim_dataset}.
We then analyze the differences of datasets in multiple aspects, and introduce our proposed domain adaptation benchmarks in \cref{sec:da_benchmark}.

\subsection{Existing indoor 3D object detection datasets}
\label{sec:existing_dataset}

Existing indoor 3D object detection approaches are primarily evaluated on ScanNet \cite{dai2017scannet} and SUN RGB-D \cite{song2015sun} datasets.
3D Front \cite{fu20213d} is also introduced as a synthetic dataset for indoor point cloud detection.
ScanNet \cite{dai2017scannet} is one of the most widely used datasets for indoor 3D scene understanding, consisting of 2.5M RGB-D video frames from real-world rooms, reconstructed into CAD models and converted into point clouds. 
As shown in \cref{fig:dataset-vis}, scenes in ScanNet dataset are fully scanned thus has a relatively high quality.
The standard split includes 1,201 training scenes and 312 testing scenes, with annotations for 529 object categories.
SUN RGB-D \cite{song2015sun} comprises 10,335 single-frame RGB-D images collected from four different sensors. 
It offers tools for point cloud generation and includes 5,073 training scenes and 4,828 test scenes, annotated with 620 object categories for 3D detection and orientation estimation.
As shown in \cref{fig:dataset-vis}, SUN RGB-D scenes have obvious point ommision areas.
3D Front \cite{fu20213d} contains over 6,800 synthetic indoor scenes with detailed 3D furniture models and designer-created layouts. 
It provides mesh models and point cloud generation tools via rendering and depth fusion, with 5,080 scenes for training and 370 for testing in semantic segmentation and layout tasks.
As shown in \cref{fig:dataset-vis}, 3D Front scenes have uniformly sampled high-quality 3D point clouds.

The scale of indoor 3D datasets is often constrained by the substantial human effort required for data collection and calibration.
Manually annotated datasets also inevitably suffer from label noise~\cite{yu20243d}.
As a synthetic dataset, 3D Front alleviates the burden of data collection. 
However, it still relies on expert-designed room configurations, which limits its scalability.
Furthermore, existing datasets lack extensibility and are insufficient for systematically investigating the various types of domain gaps that may arise in real-world applications.

\subsection{Our proposed ProcTHOR-OD and ProcFront}
\label{sec:sim_dataset}

To address the data scale issue, researches have tried to use simulators to create large-scale datasets with precise annotations at a low cost such \cite{zheng2020structured3d,fu20213d}.
However, the human designed indoor scenes still faces the high cost of construction and can hardly be scaled up.

To address the scarcity of synthetic indoor 3D object detection datasets and to enable controlled analysis of domain gap factors, we construct fully synthetic datasets using the AI2-THOR simulation platform.
Following the ProcTHOR pipeline \cite{deitke2022️procthor}, we randomly generate large-scale, diverse single-room layouts through a multi-stage process, including room specification, floorplan generation, and the placement of structural elements, large objects, wall-mounted items, and surface objects.
This procedure yields accurate object and structure information without manual annotation.
The generated layouts are exported in mesh format (e.g., .glb), preserving geometry and surface details, from which we uniformly sample points to construct point clouds.
We generate 10,000 scenes and randomly split 8,000 for training set and 2,000 for evaluation.
Our ProcTHOR-OD dataset differs from procthor-10k \cite{deitke2022️procthor} in two key ways:
(1) We generate single-room layouts instead of multi-room scenes to better suit 3D object detection instead of robotic navigation.
(2) Unlike procthor-10k, which mainly uses 2D images and floorplan configuration, we export 3D meshes and sample point clouds to support point-based learning.

\input{tables/dataset-info}

Through the benifit of scalability and extensibility, we are able to control the layout and object instances in the generated scenes to isolate the domain gap factors for a better analysis.
We construct ProcFront, a novel intermediate domain bridging ProcTHOR-OD and 3D Front, by combining the large-scale scene layouts from ProcTHOR-OD with fine-grained, category-specific 3D object models curated from the 3D Front dataset.
Specifically, with the determined object layout of ProcTHOR-OD and known instance information of 3D Front, we randomly retrieve instances in 3D Front scenes within a range of similar length, width and height.
For an instance with similar size, we place it into the origianl bounding box after size scaling and rotation to align the new instance with origianl bounding box, ensuring the correctness of annotations in the new dataset.
In this way, ProcFront share the same layout distribution with ProcTHOR and the very similar instance distribution with 3D Front to facilitate isolated analysis of domain gap factors.
The scene number of both ProcTHOR-OD and ProcFront is more than 5 times larger than ScanNet, and the number of bounding box annotations is an order of magnitude larger than ScanNet and SUN RGB-D.

\subsection{Domain adaptation benchmarks}
\label{sec:da_benchmark}

\subsubsection{Analysis of dataset status}
\label{sec:dataset_diff}

\input{figures/instance-num}

In \cref{fig:dataset-vis} we present typical scenes of each dataset.
In \cref{tab:dataset-info} and \cref{fig:instance-num} we show the statistical information of the datasets.
We analyze the differences of these datasets as follows:
(1) \textit{Data scale}: 
As shown in \cref{tab:dataset-info}, while existing real-world datasets are limited in scale and annotation quality, our proposed synthetic datasets offer low-cost, high-precision annotations with 10k scenes and 238k bounding boxes which is an order of magnitude more than ScanNet.
(2) \textit{Style(synthetic to real)}: 
As illustrated in \cref{fig:dataset-vis}, the style difference exists between synthetic and real-world datasets.
Though providing high quality point clouds, synthetic datasets lack the texture details and material realism in real-world environments.
(3) \textit{Point cloud quality}:
As shown in \cref{fig:dataset-vis}, ScanNet dataset have higher quality point clouds than SUN RGB-D which exhibits obvious point omission.
Point clouds in synthetic datasets are randomly sampled without scan omission. 
(4) \textit{Room layout}:
As shown in \cref{tab:dataset-info}, ScanNet and SUN RGB-D are collected from real-world indoor environments, featuring naturally cluttered object arrangements, whereas 3D Front consists of expert-designed scenes with clean and organized layouts. In contrast, our proposed ProcTHOR dataset adopts automatically generated layouts, which differ from human-designed ones in style but offer significantly greater scale and diversity.
(5) \textit{Instance features}:
Different datasets exhibit distinct instance-level characteristics, even within the same object category. As shown in \cref{fig:dataset-vis}, objects such as beds can vary significantly in appearance, which may mislead the detector to produce wrong prediction on categories.
Moreover, as illustrated in \cref{fig:instance-num}, the distribution of instance counts also differs notably, with our synthetic dataset providing a substantially larger number of instances.
The instance number distribution may also influence the prediction of cross-domain 3D detectors.

\input{tables/proc2scan}

\input{tables/front2scan}

\input{tables/front2sun}

\subsubsection{Domain adaptation scenarios}

Based on the analysis of differences between datasets, we propose 4 domain adaptation settings including totally 6 sets of evaluation benchmarks based on the practical requirements:

\textbf{Synthetic to real adaptation}:
Fully leveraging the large-scale, high-quality synthetic data, the detectors should be capable of adapting to real-world scenarios to meet practical requirements.
Under this setting, we use our proposed ProcTHOR-OD as source domain and ScanNet as target domain.
Following \cite{wang2024ohda}, we also evaluate 3D Front $\rightarrow$ ScanNet and 3D Front $\rightarrow$ SUN RGB-D.

\textbf{Point cloud quality adaptation}:
Detectors trained on thoroughly scanned datasets should be able to adapt to environments where the point clouds are of lower quality.
Under this setting, we use ScanNet as source domain and SUN RGB-D as target domain.

\textbf{Layout adaptation}:
Instance placement layout is a core feature of a 3D scene in detection task.
By rule-based framework, our proposed ProcFront has large scale diverse layout and at the same time shares the same instances with 3D Front.
We use ProcFront as source domain and 3D Front as target domain, hoping the detector to generalize from generated layout to manually designed layout.

\textbf{Instance adaptation}:
To evaluate instance-level generalization ability of detectors, we control the layout by ProcTHOR-OD and ProcFront, only studying the instance difference.
We use ProcTHOR-OD as source domain and ProcFront which contains additional instances as target domain.

For each domain-adaptation benchmark, we retain only the categories shared by the source and target datasets, excluding structural objects (e.g., windows, doors, pictures) and very small objects (e.g., phones, eggs, knives). 
The resulting category sets for each benchmark are listed in the tables in \cref{sec:exp_domain_gaps}.

%% file: tables/dataset-info.tex
\begin{table}
    \caption{Scale statistics of different datasets}
    \label{tab:dataset-info}
    \setlength{\tabcolsep}{2pt}
    \centering
    \small
    \begin{tabular}{ccccc}
        \toprule
        Dataset & Point cloud source & Train / val & Object annotation \\
        \midrule
        SUN RGB-D \cite{song2015sun} & Single RGB-D images & 5,073 / 4,828 & 54,376  \\
        ScanNet \cite{dai2017scannet} & RGB-D videos & 1,201 / 312 & 29,035 \\
        3D Front \cite{fu20213d} & Synthetic meshes & 5,080 / 370 & 31,022 \\
        ProcTHOR-OD & Synthetic meshes & 8,000 / 2,000 & 238,003 \\
        ProcFront & Synthetic meshes & 8,000 / 2,000 & 238,003 \\
    \bottomrule
  \end{tabular}
\end{table}

%% file: figures/instance-num.tex
\begin{figure}
    \vspace{-1mm}
    \centering
    \begin{center}
    \includegraphics[width=0.45\textwidth]{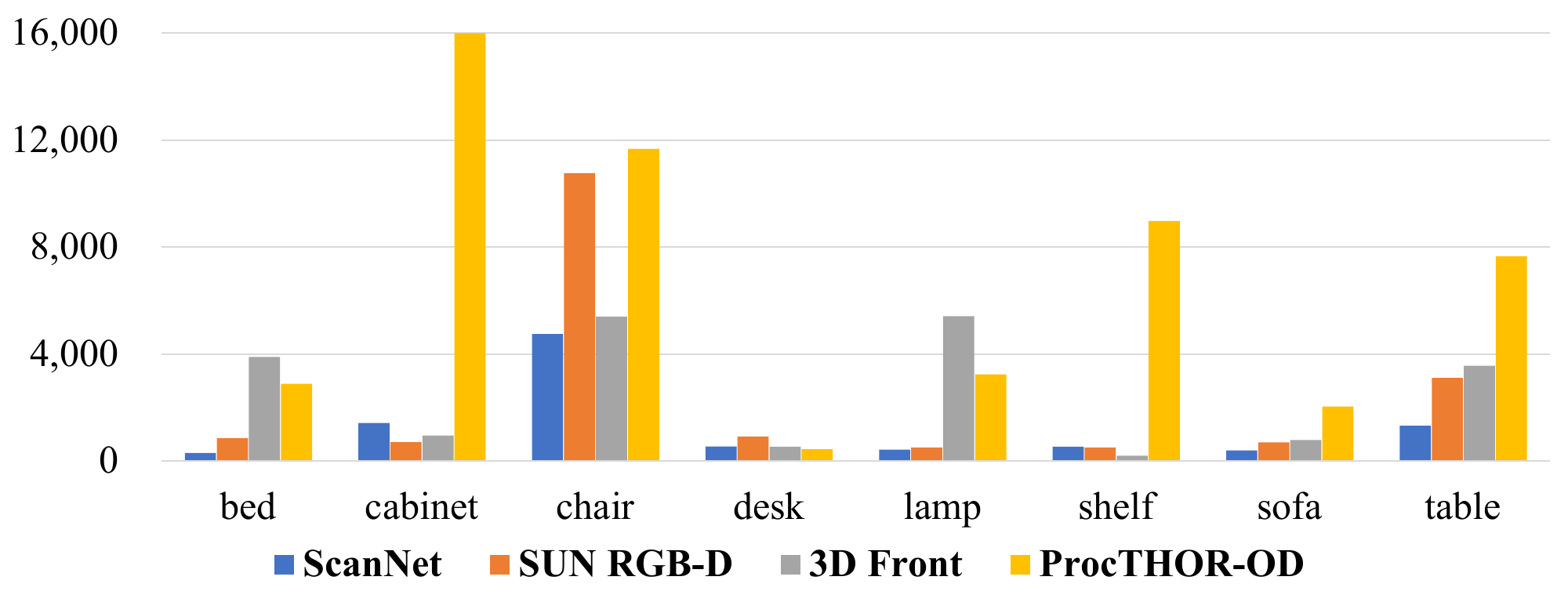}
    \end{center}
    \vspace{-3mm}
    \caption{
        Instance count by category across datasets.
    }
    \label{fig:instance-num}
    \Description{}
    \vspace{-1mm}
\end{figure}

%% file: tables/proc2scan.tex
\begin{table}
    \caption{Cross domain mAP on ProcTHOR-OD $\rightarrow$ ScanNet}
    \vspace{-1mm}
    \label{tab:proc2scan}
    \setlength{\tabcolsep}{3pt}
    \centering
    \small
    \begin{tabular}{c|cccccccc|c}
        \toprule
        Method & bed & cabinet & chair & desk & lamp & shelf & sofa & table & mAP \\
        \midrule
        source & 23.33 & 0.46 & 41.45 & 6.90 & 1.68 & 2.99 & 44.44 & 30.11 & 18.92 \\
        target & 77.69 & 22.41 & 64.91 & 55.81 & 12.68 & 19.59 & 71.79 & 45.62 & 46.31 \\
    \bottomrule
  \end{tabular}
\end{table}

%% file: tables/front2scan.tex
\begin{table}
    \vspace{-3mm}
    \caption{Cross domain mAP on 3D Front $\rightarrow$ ScanNet}
    \label{tab:front2scan}
    \setlength{\tabcolsep}{1pt}
    \centering
    \small
    \begin{tabular}{c|cccccccccc|c}
        \toprule
        Method & bed & bk.shf & cabnt & chair & desk & lamp & n.stnd & shelf & sofa & tab & mAP \\
        \midrule
        source & 12.73 & 12.65 & 0.78 & 38.92 & 6.82 & 0.55 & 1.01 & 1.03 & 35.93 & 15.38 & 12.58 \\
        target & 77.21 & 36.68 & 21.63 & 65.96 & 57.22 & 12.70 & 62.17 & 23.58 & 73.57 & 50.12 & 48.08 \\
    \bottomrule
  \end{tabular}
\end{table}

%% file: tables/front2sun.tex
\begin{table}
    \caption{Cross domain mAP on 3D Front $\rightarrow$ SUN RGB-D}
    \label{tab:front2sun}
    \setlength{\tabcolsep}{1pt}
    \centering
    \small
    \begin{tabular}{c|cccccccccc|c}
        \toprule
        Method & bed & bk.shf & cabnt & chair & desk & lamp & n.stnd & shelf & sofa & tab & mAP \\
        \midrule
        source & 31.87 & 1.52 & 0.20 & 25.20 & 4.91 & 0.17 & 0.24 & 0.12 & 13.14 & 21.29 & 9.87 \\
        target & 85.33 & 33.74 & 8.52 & 71.54 & 26.94 & 32.00 & 65.29 & 12.87 & 65.70 & 53.33 & 45.52 \\
    \bottomrule
  \end{tabular}
\end{table}

%% file: chapters/4.Experiment.tex
\section{Experiments and analysis}

\input{tables/scan2sun}

\input{tables/pf2front}

\input{tables/proc2pf}

\subsection{Setup and details}

In our experiments, we apply commonly used indoor 3D object detector VoteNet \cite{qi2019deep} to evaluate the cross domain performance and domain adaptation approaches.
We randomly sample 40,000 points per scene for training.
In addition to coordinates, we include height and RGB color features for each point.
We train the network end-to-end with AdamW optimizer and batch size 32 for 90 epochs.

During inference, we post-process the proposals with dropping empty boxes (less than 5 points in a box), dropping too little boxes (boxes with side length less than 0.001 m) and 3D NMS (non-maximum suppression) with IoU threshold of 0.25.
We report the evaluation metric of mAP (mean average precision) following the protocal of \cite{song2016deep} with IoU threshold 0.25.
More details can be found in our open source code.

\subsection{Analysis of domain gaps}
\label{sec:exp_domain_gaps}

We present the basic experimental results on our proposed benchmarks in \cref{tab:proc2scan} to \ref{tab:proc2pf} and further domain adaptation results in \cref{tab:baseline-results} to analysis the impact of multiple domain gap factors mentioned in \cref{sec:dataset_diff}.

\textbf{Style(synthetic to real)}:
ProcTHOR-OD $\rightarrow$ ScanNet, 3D Front $\rightarrow$ ScanNet and 3D Front $\rightarrow$ SUN RGB-D exhibit synthetic to real style gap, as illustrated in \cref{tab:proc2scan}, \cref{tab:front2scan} and \cref{tab:front2sun}.
ScanNet trained detector performs well on its own evaluation set, but drastically decline to only 12.58 mAP when trained on 3D Front.
The result on ProcTHOR trained model is slightly better (18.92 mAP), illustrating that data scale contributes to performance even when domain gap is rather large.
SUN RGB-D trained detector has 45.52 mAP within its domain, but declines to only 9.87 mAP when trained on synthetic dataset.
The SUN RGB-D serve as a more challenging target domain dataset, given the low quality point cloud that is converted from single RGD-B images.
Experiments shows that synthetic-to-real gap is a challenging hurdle for domain adaptive indoor 3D object detection among our proposed benchmarks.

\textbf{Point cloud quality}:
The ScanNet$\rightarrow$SUN RGB-D benchmark isolates the point-cloud quality gap: adapting between two real datasets avoids confounds from instance shape and texture. 
As shown in \cref{tab:scan2sun}, a model trained on ScanNet attains 23.56 mAP on SUN RGB-D—well below the 45.70 mAP within domain—highlighting the risk of deploying models trained on well-scanned data to lower-quality point clouds and the value of this benchmark.

\textbf{Layout adaptation}:
As shown in \Cref{tab:pf2front}, the flexibility of our generative data framework lets us isolate the effect of the layout domain gap. 
Although ProcFront uses the same instance models as 3D-Front, switching from automatically generated layouts to human-designed ones sharply reduces performance—from 93.81 to 22.62 mAP. 
This highlights scene layout as a major driver of domain shift in object detection.

\textbf{Instance adaptation}:
As the fundamental task of object detection, the performance not only requires localization, but also recognition of instances.
In \cref{tab:proc2pf} we control the layout as the same and evaluate the impact of instance change.
Results shows that even with exactly the same layout, instance feature gap also cause a decline in performance, from 73.28 to 25.64 mAP.

\input{tables/da-approach}

\subsection{Domain adaptation approaches}
\label{sec:exp_da_approaches}

As shown in \cref{tab:baseline-results},  we introduce several domain adaptation approaches including methods using target domain priors (block ``Prior'' in the table) and unsupervised domain adaptation techniques (block ``UDA'' in the table) to improve adaptation performance, presenting a first series of baselines for this problem.

\subsubsection{Using target domain priors}

When the domain gap is substantial, lightweight target-domain priors—far easier to obtain than exhaustive target annotations—can significantly boost cross-domain detection performance. 
In the ``Prior'' block of \cref{tab:baseline-results}, we report results for different types of priors. 

\textbf{Using object size prior:}
Different datasets exhibit average object size difference.
By simply adopting the target domain statistical prior of mean sizes (instead of precise labels) for each category, reported in line ``Size'', the detector shows performance gains across all the benchmarks. 
This observation shed a light on future studies to estimate target domain object sizes.

\textbf{Few-shot fine-tuning:}
In line ``10 shots'' and ``100 shots'' we report the result of fine-tuning the source-only trained model with 10 and 100 labeled target scenes.
On all the baselines, 10-shot fine-tuning improves the performance only slightly relative to the source-only baseline. 
Using 100 target domain annotated scenes yields substantial performance gains.
However, performance still lags far behind the target-domain oracle. 
Given the high cost of collecting and annotating real-world target scenes, unsupervised domain adaptation remains a highly desirable direction.

\subsubsection{Unsupervised Domain Adaptation}

Unsupervised domain adaptation setting requires only the point clouds without the need of annotation.
We evaluate different domain adaptation techniques on our proposed benchmarks, including vistual scan simulation \cite{ding2022doda} (reported in line ``VSS''), mean teacher (reported in line ``MT'') \cite{tarvainen2017mean}, and reliable voting \cite{fan2022self} (reported in line ``RV'').
Note that due to the difference on task setting, we re-implement all the methods and make minor adjustments for object detection task.

\textbf{Virtual scan simulation.}
Introduced by DODA \cite{ding2022doda} for domain-adaptive indoor point cloud segmentation, Virtual Scan Simulation (VSS) targets the synthetic-to-real gap by mimicking imperfections in real reconstructions. 
VSS models (i) sensor noise, perturbing each point’s $(x,y,z)$ coordinates, and (ii) view occlusion, removing occluded points via multi-view projection.

\textbf{Mean teacher.}
Originally proposed for semi-supervised learning \cite{tarvainen2017mean} and now widely used for UDA across classification, detection, and segmentation on both 2D images \cite{zhao2023masked} and 3D LiDAR point clouds \cite{chang2024cmda}, the mean-teacher framework trains a source-only model and initializes a teacher that generates pseudo-labels for target samples to supervise a student; the teacher is typically updated from the student (e.g., via EMA) to enforce consistency.

\textbf{Reliable voting.}
Proposed for domain-adaptive point cloud classification \cite{fan2022self}, reliable voting improves pseudo label selection in the mean-teacher framework. 
In our implementation, we cache features and predicted categories for all source-domain instance proposals; during training, for each target proposal we retrieve its top-$K$ most similar source proposals and check whether their categories agree with the target prediction. 
We accept the target pseudo-label only when this cross-domain vote is reliable.

As shown in the ``UDA'' block of \cref{tab:baseline-results}, VSS augmentation performs well in synthetic-to-real settings but lacks generality for layout and instance adaptation.
(The augmented data has lower point cloud quality, thus harms the performance where the target domain has high-quality point clouds in ``pf2front'' and ``proc2pf''.)
The mean-teacher approach consistently works across all baselines, demonstrating its generality, and reliable voting (RV) consistently improves mean-teacher results, indicating the effectiveness of pseudo-label filtering.
More details can be seen in our open-source code.

%% file: tables/scan2sun.tex
\begin{table}
    \caption{Cross domain mAP on ScanNet $\rightarrow$ SUN RGB-D}
    \label{tab:scan2sun}
    \setlength{\tabcolsep}{0.8pt}
    \centering
    \footnotesize
    \begin{tabular}{c|cccccccccccccc|c}
        \toprule
        Method & bed & bk.shf & cbnt & chr & desk & gbg.c & lamp & nstnd & shf & sink & sofa & tab & tlt & tv & mAP \\
        \midrule
        source & 58.7 & 12.3 & 1.9 & 38.5 & 11.7 & 15.4 & 7.6 & 20.1 & 6.5 & 16.5 & 42.3 & 32.9 & 56.8 & 8.5 & 23.6 \\
        target & 82.1 & 28.7 & 8.3 & 71.5 & 24.7 & 43.9 & 33.0 & 62.2 & 13.5 & 51.2 & 63.8 & 50.6 & 86.2 & 20.0 & 45.7 \\
    \bottomrule
  \end{tabular}
\end{table}

%% file: tables/pf2front.tex
\begin{table}
    \caption{Cross domain mAP on ProcFront $\rightarrow$ 3D Front}
    \label{tab:pf2front}
    \setlength{\tabcolsep}{2pt}
    \centering
    \small
    \begin{tabular}{c|ccccccccc|c}
        \toprule
        Method & bed & cabnt & chair & desk & lamp & shelf & sofa & tab & tv.stnd & mAP \\
        \midrule
        source & 95.53 & 0.87 & 59.54 & 15.45 & 0.28 & 0.34 & 41.86 & 38.80 & 21.11 & 30.42 \\
        target & 99.99 & 74.73 & 97.70 & 83.03 & 95.52 & 72.52 & 99.27 & 92.28 & 0.63 & 79.52 \\
    \bottomrule
  \end{tabular}
\end{table}

%% file: tables/proc2pf.tex
\begin{table}
    \caption{Cross domain mAP on ProcTHOR-OD $\rightarrow$ ProcFront}
    \label{tab:proc2pf}
    \setlength{\tabcolsep}{2pt}
    \centering
    \small
    \begin{tabular}{c|ccccccccc|c}
        \toprule
        Method & bed & cabnt & chair & desk & lamp & shelf & sofa & tab & tv.stnd & mAP \\
        \midrule
        source & 33.13 & 0.23 & 42.35 & 3.59 & 3.76 & 2.56 & 63.28 & 46.11 & 38.22 & 25.92 \\
        target & 85.59 & 21.32 & 65.26 & 56.23 & 35.99 & 10.58 & 87.13 & 73.37 & 56.68 & 54.68 \\
    \bottomrule
  \end{tabular}
\end{table}

%% file: tables/da-approach.tex
\begin{table}
    \caption{Baseline results on our proposed benchmarks}
    \label{tab:baseline-results}
    \centering
    \small
    \setlength{\tabcolsep}{1pt}
    \begin{tabular}{cccccccc}
        \toprule
        & Method & proc2scan & front2scan & front2sun & scan2sun & pf2front & proc2pf \\
        \midrule
        & Source & 18.90 & 12.58 & 9.87 & 23.56 & 30.42 & 25.92 \\
        \midrule
        \multirow{3}{*}{Prior} & Size & 18.92 & 13.53 & 9.92 & 25.56 & 32.43 & 28.27 \\
        & 10 shots & 17.47 & 14.05 & 10.39 & 24.05 & 32.56 & 27.00 \\
        & 100 shots & 33.86 & 25.83 & 10.54 & 24.58 & 37.01 & 29.23 \\
        \midrule
        \multirow{3}{*}{UDA} & VSS \cite{ding2022doda} & 20.31 & \textbf{16.72} & \textbf{10.61} & \textbf{25.26} & 28.15 & 24.32 \\
        & MT \cite{tarvainen2017mean} & 20.43 & 15.03 & 10.32 & 25.25 & 32.70 & 27.44 \\
        & RV \cite{fan2022self} & \textbf{20.89} & 15.27 & 10.40 & 24.92 & \textbf{32.74} & \textbf{27.78} \\
        \midrule
        & Oracle & 46.31 & 48.08 & 45.53 & 45.70 & 79.52 & 54.68 \\
    \bottomrule
  \end{tabular}
\end{table}

%% file: chapters/5.conclusion.tex
\section{Conclusion}

In this paper, we conduct a comprehensive analysis of domain gaps across indoor 3D object detection datasets. 
By combining three widely used datasets: ScanNet, SUN RGB-D, and 3D-Front, with our two newly proposed datasets, ProcTHOR-OD and ProcFront, we build the first comprehensive benchmark suite for domain adaptive indoor 3D object detection. 
This suite comprises six benchmarks spanning four adaptation scenarios: synthetic-to-real, point cloud quality, layout, and instance adaptation. 
We introduce several baseline methods and evaluate them on our benchmarks, highlighting future directions for improving adaptation performance.

\begin{acks}
This work was supported by the
grants from the National Natural Science Foundation of
China (62372014, 62525201, 62132001, 62432001), Beijing Nova Program and Beijing Natural Science Foundation (4252040, L247006).
\end{acks}